# BD-SAT: High-resolution Land Use Land Cover Dataset & Benchmark Results for Developing Division: Dhaka, BD

Ovi Paul†, Abu Bakar Siddik Nayem†, Anis Sarker, Amin Ahsan Ali, M Ashraful Amin, and AKM Mahbubur Rahman

Independent University Bangladesh
Center for Computational & Data Sciences
Dhaka, Bangladesh

†These authors contributed equally to this work

## ABSTRACT

Land Use Land Cover(LULC) analysis on satellite-images using deep learning-based methods are significantly helpful in understanding target area's geography, socio-economic condition, poverty-level, and urban-sprawl in developing countries. Plenty of recent works involve segmentation with LULC classes like farmland, built-up, forest, meadow, water, etc. Piles of images with LULC-annotated classes are required to train deep-learning methods on satellite-images. However, annotated data for developing countries are scarce for lack of funding; lack of dedicated residential/industrial/economic zones; huge population; and diverse building materials. BD-SAT provides high-resolution dataset that includes pixel-by-pixel LULC annotations for Dhaka metropolitan city and surrounding rural/urban area. With strict and standard procedure, the ground-truth is made using Bing-satellite imagery with ground-spatial-distance of 2.22 meters/pixel. Three stages well-defined annotation process has been followed with support from GIS experts to ensure the reliability of the annotations. We perform several experiments to establish the benchmark results. Results show that the annotated BD-SAT is sufficient to train large deep-learning models with adequate accuracy for five major LULC classes: forest, farmland, built-up, water, and meadow.

## 1 Introduction

Land use/land cover(LULC) changes have gained major interest as the ecosystem changes rapidly at a global scale[1]. Most importantly, developing countries are experiencing substantial LULC changes across the world[2,3]. In recent decades, grassland, woodland, bushland and other vegetation covers have been extensively transformed into agricultural/settlement areas to accommodate the enormous population increment and their demands[4,5]. Land cover is defined as the physical material covering the earth's surface. Forest, mountains, desert, and water are examples of land cover[6]. Land use is the description of how humans utilize the land for socio-economic activities; i.e, farmland, built-up, etc.[7,8]. Thus, data on the change of LULC over a period can be used to understand the geography, socio-economic condition, poverty-level, and urban-sprawl of an area[9,10,11,12]. Recently, deep-learning based methods have proved to be effective to extract LULC information from satellite-images of a vast regions and have been widely used in the above applications[13,14,15,16,17,18]. However, we must address two issues to use deep-learning methods effectively to perform LULC on developing countries. The first issue is the lack of LULC-annotated data from developing countries. We need to have large amounts of images annotated with ground-truth LULC classes. Particularly, pixel-by-pixel LULC class annotation is essential to train deep learning methods for the applications mentioned above. Nonetheless, annotated and representative data availability becomes crucial for developing countries. A large number of research groups produce annotated satellite data with huge financial supports[19,20,21,22,23] in developed countries. However, they might not be useful since training deep-learning models with annotated datasets collected in developed countries yields erroneous results when applied on developing countries. Moreover, urban structures across different countries vary significantly due to building materials, land characteristics, and cultures[24]. Hence, LULC annotations with high-resolution satellite data from developing countries are of the utmost necessity. Due to the lack of funds, very few datasets are annotated for developing countries (e.g. Bangladesh, India, Srilanka, African countries[25,26,5,27,24,15]; table 1).

**Table 1.** Summary of the LULC datasets from developing countries

| Dataset Name | Region | Purpose | Source | Time of Acquisition | GSD (Per Pixel) |
|---|---|---|---|---|---|
| In-Sat[25], 2021 | India | LULC | Sentinel-2 | Jan - April, 2020 | 10 m |
| IndiaSat[26], 2021 | India | LULC | Landsat-7, Landsat-8 and Sentinel-2 | 2016-2019 | 10, 15 m |
| Mpologoma Catchment[5], 2021 | Eastern Uganda | LULC | Landsat 4-5 TM, Landsat 7 ETM+ and Landsat 8 OLI/TIRS | July - November, 2019 | 15 m |
| Demissie et al.[15], 2017 | Ethiopia | LULC | Landsat MSS, TM,ETM+ and Landsat 8 OLI | 1973, 1985, 1995, 2003 and 2015 | 15 m |
| LandCoverNet[28], 2020 | Africa, Asia, Australia, Europe,North America and South America | LULC | Sentinel-2 | 2018 | 10 m |
| DeepGlobe[29], 2018 | Thailand, Indonesia and India | Road Extraction | WorldView 2-3 | 2009, 2011, 2012 and 2013 | 10cm and 8cm |
| JECAM[30], 2021 | Africa | Agriculture | Pleiades and SPOT-6 | 2013-2021 | 0.25m and 1.8m |
| Peng et al.[31], 2020 | Africa | Drought | SPEI | 1981-2016 | 5km |
| World bank[27], 2021 | Dhaka | LULC | QuickBird, Pleiades, Landsat and Sentinel-2 | 2006,2017 | 0.5m-30m |
| **BD-SAT(Ours)** | **Dhaka** | **LULC** | **Bing** | **2019** | **2.22 m** |

Specifically, World Bank has developed an urban mapping of Dhaka metropolitan city[27], but it doesn't have the important details (See fig. 1) for pixel-by-pixel annotations. Datasets collected in India[25,26] do not have good resolutions. They have used patches to annotate into four classes: Forest/Water/Urban/others. Low-resolution and patch-based techniques induce errors for the deep learning models applied in highly-populated and unstructured developing countries. Because, unlike developed countries, developing cities do not have dedicated residential/industrial areas and pre-planned marketplaces. Table 1 provides the summary of the available LULC datasets collected from developing countries.

Secondly, most of the deep-learning methods use only RGB channels([32]) for LULC on satellite images and achieve superior performance compared to the traditional methods[33,14]. However, high-resolution RGB data might not be available frequently for applications. Though Bing/Google provides high-resolution RGB data, they don't update their freely available images regularly for developing countries[34,35,36,37]. Highly populated western regions get frequent updates; on the contrary, developing countries like Bangladesh get updates at less frequently because they generate comparatively lower profits. Satellite data (e.g., sentinel-2/sentinel-3/Landsat) are available frequently (weekly/biweekly/monthly). Moreover, they have several sensing channels. The remote-sensing researchers traditionally use various indices (e.g., normalized vegetation index-NDVI, normalized moisture index-NDWI, radar vegetation index-RVI) computed from different channels to identify LULC information[38,39,40]. Therefore, using different channels in deep learning should be beneficial despite satellite data (Sentinel-2/sentinel-3) having much lower resolution than Bing/Google imagery.

Contributions of this paper are summarized here. **At first**, a new dataset has been introduced in this paper. BD-SAT is annotated with the highest care and immaculate details. BD-SAT includes the Dhaka metropolitan city and the surrounding rural/urban area. After a long and strict annotation process, BD-SAT provides a highly accurate ground truth for a developing country's urban/rural regions. The ground truth was made using Bing satellite imagery at 17 zoom-level (ground-spatial distance of 2.22 meters/pixel). Therefore, BD-SAT has annotations with very high-resolution images, so it might be helpful for any developing countries in South-Asia/East Asia (Indonesia, Malaysia, India, Srilanka) with similar topography.

**In second**, we perform several experiments to benchmark the performance of the state-of-the-art deep-learning algorithm: DeepLabV3+[41] on Sentinel-2A data (low-resolution) and Bing data (high-resolution). Specifically, data from different channels from Sentinel-2A are used with the same BD-SAT ground truth. Experiments on combinations (AtmosphericPenetration-ATM; FalseColorInfrared-FCI; ShortwaveInfrared-SWIR) and index images (NDVI, NDWI, RVI) have been performed to see the model performance on low-resolution images (10-20 meter/pixels). Then, we use Bing images to report the benchmark performance on high-resolution data. **Finally**, we provide conclusive recommendations regarding the scope of the BD-SAT dataset for sentinel (low-resolution data)/Bing data (any high-resolution data i.e, Google) to be used with the deep-learning models for developing countries.

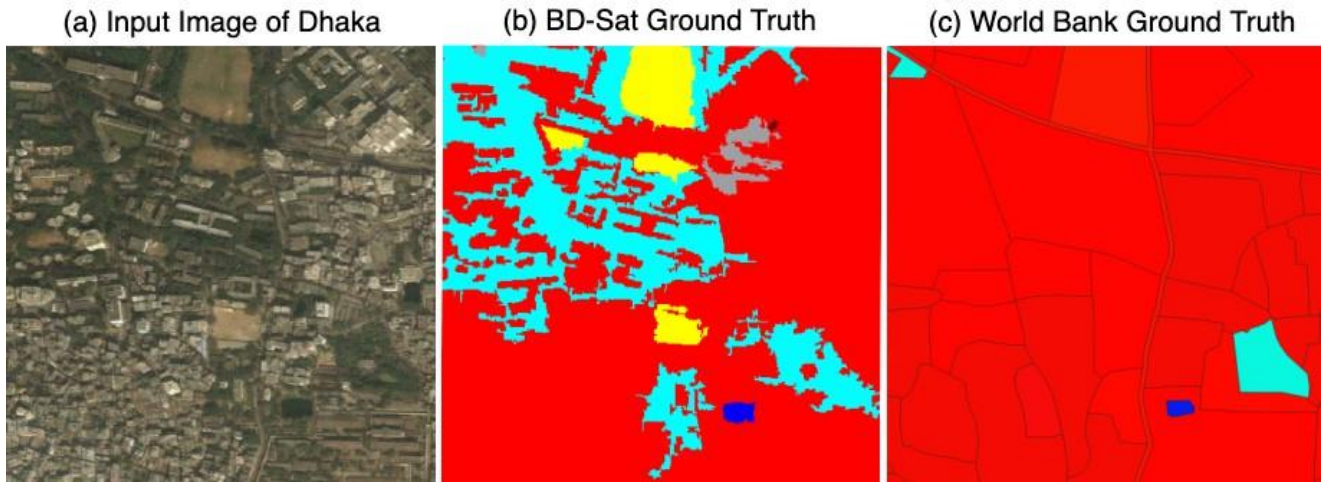

**Figure 1.** Comparison between World Bank and BD-SAT(ours) Ground Truth (Here red color represents various kinds of built-up, cyan represents forest/tree cover, yellow represents bare land/meadow and blue represents water)

## 2 Literature Review

In this section, we provide the background studies and the context for the LULC with satellite images. We first describe the existing satellite data that are pixel by pixel annotated. Secondly, we summarize state-of-the-art deep models used for segmenting satellite images. Then we discuss the sentinel data channels, their characteristics, usage, and information derived from them.

**LULC datasets**: There are many LULC datasets with good-quality annotations. Most of them are from Europe. The popular one is the Potsdam and the Vaihingen dataset, which covers the areas in Germany[19]. The classes are building, low-vegetation, tree, car, and background. Gaofen dataset (GID) is based on different regions in China. It has five classes- forest, built-up, farmland, meadow, and water[22]. Another dataset suitable for land-cover semantic segmentation is LoveDa[23]. LoveDa stands for Land-cOVEr Domain Adaptive, which contains high resolution from urban scenes to rural scenes. The LoveDA dataset was constructed using 0.3 m images from Nanjing, Changzhou, and Wuhan. MiniFrance[21] is another dataset similar to LoveDa. As the name suggests, MiniFrance data covers the regions in France. It gives annotation to the urban and rural scenes. But MiniFrance data is aimed at semi-supervised learning. FloodNet[20] is a LULC dataset with ten classes, one of which is background. As the name suggests, the FloodNet dataset deals with flood-related data. The dataset was collected during a flood in Texas and Louisiana in the USA. India has In-Sat dataset from developing regions that sourced images from Sentinel-2 data and Google Earth data. The dataset is made to control the environment, urban planning, and land cover classification[25]. Similarly, IndiaSat is another dataset for land cover classification based on different geographic regions in India. The dataset consists of 180,414 pixels labeled into four land cover classes, e.g., greenery, water bodies, barren land, and built-up area. The data were sourced from Landsat-7, Landsat-8, and Sentinel-2[26]. In the dataset for Mpologoma catchment[5], a region in eastern Uganda. The study used both ancillary and satellite data. Ancillary data included aerial images and ground truth data. All the data were collected from July to November 2019 from Landsat 4-5 TM, Landsat 7 ETM+, and Landsat 8 OLI/TIRS. Another African dataset is in Ethiopia[15]. The study was conducted in the Libokemkem district in South Gondar, which covers 1082 square kilometers. The images for this study was taken from Landsat MSS, TM, ETM+, and Landsat 8OLI of the year 1973, 1985, 1995, 2003, and 2015. Satellite images from various regions in China were accumulated and annotated in the Gaofen Image Dataset[22]. Traditionally, the GIS community produces annotations using QGIS/ ArcGIS with coarse boundaries. This software creates erroneous annotation where many pixels are incorrectly annotated inside the coarse boundary though those pixels belong to different classes. These errors are prominent in developing countries, e.g., Bangladesh.

For example, World Bank has developed an urban mapping of Dhaka[27], but it lacks intricate details. It involves categories such as construction sites, airports, urban and industrial units indeterminable from imagery only. Existing datasets, e.g., world bank[27] data, suffer several limitations: a) it has a coarse level of annotations where some important classes are merged, b) only urban area is annotated, and c) it is not free. The limitations are described here. Firstly, several important classes are merged. For example, the annotation data of the World Bank do not contain any vegetation/tree cover, as shown in 1. The cyan-colored region in our ground truth represents forest/tree cover, but in World Bank ground truth, the built-up and vegetation areas are put in a single class. The separation between the vegetation and built-up types is needed as both classes are heavily studied to understand environmental and economic aspects. This is even more problematic for a country like Bangladesh, one of the most

3/26

densely populated countries. Here, the artificial structures, i.e., houses, industries, shopping malls, and markets, are closely knitted with natural components like rivers, farmland, forests, etc. These errors can impact deep learning models significantly as they are trained using pixel-by-pixel ground truth. Secondly, the World Bank dataset is minimal for urban environments only. No rural area is annotated as the dataset has been developed over the Dhaka metropolitan city. The image data is purchased from the respective satellite vendors (QuickBird, Pleiades, Landsat, and sentinel).

**Deep Models**: The deep learning architecture used is the DeepLabV3+ developed by Google[41]. This is the improved version of the earlier DeepLabV3 with the extension of an encoder-decoder structure. One of the main features in DeepLabV3+ is the Atrous convolution which was introduced in its previous version DeepLabV3[42]. None of the earlier studies on this architecture has been used on LULC, so our research will give new findings on its performance for LULC on remote sensing images. Even though DeepLabV3+ is a state-of-the-art deep learning model, it still faces complications working with the class-imbalanced dataset. In our dataset, few classes, like water, encompass proportionally smaller regions than other classes. To solve this issue, the current loss function of DeepLabV3+ was removed, and a Balanced Cross Entropy was added instead. This loss function will emphasize more the error coming from smaller classes in the data set. As a result, those classes will perform better[43]. The image was divided into square patches and exploited the context information from neighboring patches using the spatial attention module to segment the targeted patch[44]. This was implemented in the network to segment high-resolution images more significantly. Modified Unet[45] has been used to classify land use as a land cover on the pixel level where Garg et al. have reduced the number of layers from 23 to 19, increased the filter size from 3x3 to 5x5, and used padded convolutions contrary to the original Unet[46]. The efficiency of various deep learning architectures for LULC classification is measured by Papoustsis et al.[47], where authors depict that Resnet50[48] and VGG19[49] perform better than others while classifying LULC patches from the BigEarthNet[50] dataset. Sathyanarayanan et al.[51] have modified segnet[52], which then was trained on five spectral bands, namely Blue, Green, Red, Near-Infrared (NIR), and Short wave Infrared (SWIR) from the Sentinel-2A satellite, covering Mandya, Karnataka, India and predicted for other similar regions in India and found promising results. Zhang et al.[53] proposed Joint Deep Learning (JDL) for LULC classification, incorporating patch-based CNN and pixel-based MLP with joint reinforcement and mutually complementary. Here, the joint distributions between LC and LU were formulated into a Markov process, which improves accuracy through iterative updating.

**Sentinel Data Description**: From satellites, combination images are used, which consist of multiple bands. Combination imagery like Short Wave Infrared(SWIR) is used for moisture content[54], mineral distinctions, False Color Infrared(FCI) is used for vegetation detection, and Atmospheric Penetration emphasizes classes similar to that of traditional false-color infrared photography but has an excellent clarity[55]. The RGB satellite image was also combined with experimenting and compared with other combinations and indices. Three types of index images were used for our experiments. First is the Normalized difference vegetation index(NDVI). Researchers can calculate the escalation of light coming off the soil in noticeable and near-infrared wavelengths and evaluate the photosynthetic capacity of the vegetation in a given pixel. More reflected radiation in near-infrared than in detectable wavelengths; at that point, the vegetation in that pixel is likely to be thick and may contain some woodland. The minimal distinction within the concentration of specific and near-infrared wavelengths reflected; at that point, the vegetation is likely undesirable or scanty and may comprise prairie, tundra, or desert. This vegetation index is used for climate change monitoring, drought monitoring, vegetation monitoring, crop evaluation and etc[56,57]. The radar Vegetation Index(RVI) works similarly to NDVI. RVI is used for estimating the Vegetation Water Content of Rice and soybeans. RVI and NDVI are vegetation indexes that detect vegetation using different formulas[58].

The NDWI was derived using principles similar to those that were used to derive the NDVI. The green band in NDWI encompasses reflected green light, and Near-infrared(NIR) represents reflected near-infrared radiation. The selection of these wavelengths was made to (i) maximize the typical reflectance of water features by using green light wavelengths; (ii) minimize the low reflectance of NIR by water features; and (iii) take advantage of the high reflectance of NIR by terrestrial vegetation and soil features. Soil and earthly vegetation features have zero or negative values, owing to their higher reflectance of NIR than greenlight.

Image handling programs can effectively be arranged to erase negative values. This eliminates the earthbound vegetation and soil data and holds the open-water data for analysis. The extent of NDWI is at that point from zero to one. Index images are being used to detect vegetation, water and etc[59]. These classes are separated in the image by manually setting threshold values. The issue with classifying a threshold value is that the same threshold does not apply to all regions. In our experiments with Dhaka's sentinel-2 tile, we used reference values to classify the Dhaka city region. However, the performance was not good when comparing the segmented image using the threshold value against the ground truth of Dhaka. In many areas, two different classes, farmland, and water, could not be separated using a threshold value. But our deep learning model, it takes more features of the images, not just the reflectance value like the existing methods. Our model learns the local geography of the region and can classify the areas with high accuracy. In the end, a comparison between the satellite image and aerial image of Bing has been made, which is available at a higher resolution. The comparison shows the cost and benefit of using a high-resolution aerial image like Bing.

Table 2. Sentinel-2 Band information and converted spatial resolution

| Band Number | Band Name | Original Spatial Resolution (m) | Converted Spatial Resolution (m) |
|---|---|---|---|
| Band 1 | Coastal aerosol | 60 | 10 |
| Band 2 | Blue | 10 | 10 |
| Band 3 | Green | 10 | 10 |
| Band 4 | Red | 10 | 10 |
| Band 5 | Vegetation red edge | 20 | 10 |
| Band 6 | Vegetation red edge | 20 | 10 |
| Band 7 | Vegetation red edge | 20 | 10 |
| Band 8 | NIR | 10 | 10 |
| Band 8A | Narrow NIR | 20 | 10 |
| Band 9 | Water Vapour | 60 | 10 |
| Band 10 | SWIR Cirrus | 60 | 10 |
| Band 11 | SWIR | 20 | 10 |
| Band 12 | SWIR | 20 | 10 |

## 3 Dataset Collection and Annotation Procedures

In this section, we describe the source of the satellite data, the location of the data collection, and data characteristics. We select our study area which includes the Dhaka division while keeping Dhaka city in the center. The location of the image tile was - N24°23'04", E89°57'26" (Upper Left Corner) and N23°22'13", E91°00'26" (Lower Right). Therefore the study area forms a rectangle with a horizontal width of 117 km and a vertical height of 124 km.

### 3.1 Sources of satellite data

Two different sources of images are used in this research. Since Bing image has a resolution of 2.22 meter/pixel ground spatial distance, we use it for ground truth annotation. Later, experiments with high-resolution semantic segmentation have been also performed on Bing data. On the other hand, Sentinel-2A images are used for experiments only where different band combinations and index images are used as input to the Deeplabv3+ model.

#### 3.1.1 Image Details for Bing

Bing provides high-resolution aerial images of 2.22 meters/pixel ( 2.22 meters ground spatial distance) at 17 zoom levels from Bing satellite[60]. The acquisition date of the Bing image is April 20, 2019. The Bing image tile is of resolution: 48906x47256, and the tile provides an RGB image in tiff format. The tiff format allows embedding the geolocations for each pixel. In this work, we have annotated the entire Dhaka city and its surrounding area. The whole Bing image can be seen in Figure 2a where the blue rectangle shows the place where we have performed annotation. The red rectangle indicates the Dhaka city region used for testing.

#### 3.1.2 Image downloads for Sentinel 2A

The Acquisition start date was 2019-04-11T04:38:29.547Z. We have also collected Sentinel-2A data with tile number "T45QZG" and Tile ID "L1C_T45QZG_A016175_20190411T043829". The Sentinel 2A data includes information from 12 bands shown in the table 2. It can be noticed from the table 2 that data for different bands have been captured with various spatial resolutions from 10 to 60 meters/pixel. The data for red (band 2), green (band 3), blue (band 4), and near-infrared (NIR-band 2) bands are captured with 10 meter/pixel resolution, whereas the data for Coastal aerosol (band 1), Water Vapour (Band 9), and SWIR Cirrus (Band 10) are captured with 60 meter/pixel.

### 3.2 Dataset characteristics

The target is to annotate a portion (the blue rectangle) of the Bing data. There are primarily eleven classes of interest during the annotation process. The second column of Table 3 shows the class names. The fourth column shows the definition of each class. The distribution of the pixels for eleven classes is shown in fig 9a.

### 3.3 Ground Truth Annotation Process

As the Bing image has the best visual information available from 2.22 meters/pixel, we have used the Bing image for annotation so that the annotators can see the textures and colors of the lands while providing the LULC label. It is impossible to annotate the huge 48906x47256 Bing image visually. Therefore, we divide the whole Bing image by 33 X 33 grids into 1089 sub-images.

**Table 3.** Dataset Description Table

| Index Value | Original Class | RGB Values | Description | Converted Class | RGB Values |
|---|---|---|---|---|---|
| 0 | Unrecognized | 0, 0, 0 | Any part of the image that cannot be recognized as the above classes would be classified as Unrecognized. So, we don't know what class this is. | Unrecognized | 0, 0, 0 |
| 1 | Farmland | 0, 255, 0 | The segment that is used for farming. It also includes Farmland after crop harvesting and new crop plantation. Please check the context too to mark Farmland. | Farmland | 0, 255, 0 |
| 2 | Water | 0, 0, 255 | Segment that has River, Pond, and Lake, any water body. | Water | 0, 0, 255 |
| 3 | Forest | 0, 255, 255 | Segment with Single trees, a Bunch of trees, small bushes, small forest, big forest. | Forest | 0, 255, 255 |
| 4 | Urban Structure | 128, 0, 0 | When the segment is in the city areas, and both Urban Built-up and roads are present in that same segment; you can mark it as Urban Structure Built-up + Road. | Built-Up | 255, 0, 0 |
| 5 | Rural Built-Up | 255, 0, 255 | Man Made structures in village areas like houses, shops, etc. While selecting this segment for this class, a small portion of trees or Meadow might be there. If the segment mainly contains houses along some small trees/meadow portion, that is alright to name this segment Rural Built-up. | Built-Up | 255, 0, 0 |
| 6 | Urban Built-Up | 255, 0, 0 | Man Made structures in City areas. Buildings, shops in the city, factories, shopping centers, roofs, slums, structures with corrugated tin, polythene roofs, and bamboo roofs in the city area. | Built-Up | 255, 0, 0 |
| 7 | Road | 160, 160, 164 | Paths used for vehicles to move. | Built-Up | 255, 0, 0 |
| 8 | Meadow | 255, 255, 0 | Grassland or any field free of Farmland. Non-concrete rural paths are also annotated as meadow as they have soil textures. | Meadow | 255, 255, 0 |
| 9 | Marshland | 255, 251, 240 | Segment that contains Water along with either Farmland or Meadow in a single segmented piece. The segment will be considered Marshland; you can mark it as Urban Structure Built-up + Road. | Meadow | 255, 255, 0 |
| 10 | Brick Factory | 128, 0, 128 | To classify brick factories, e.g., the structure where it's made along with all the bricks that are laid down on the field, including all the paths. | Built-Up | 255, 0, 0 |

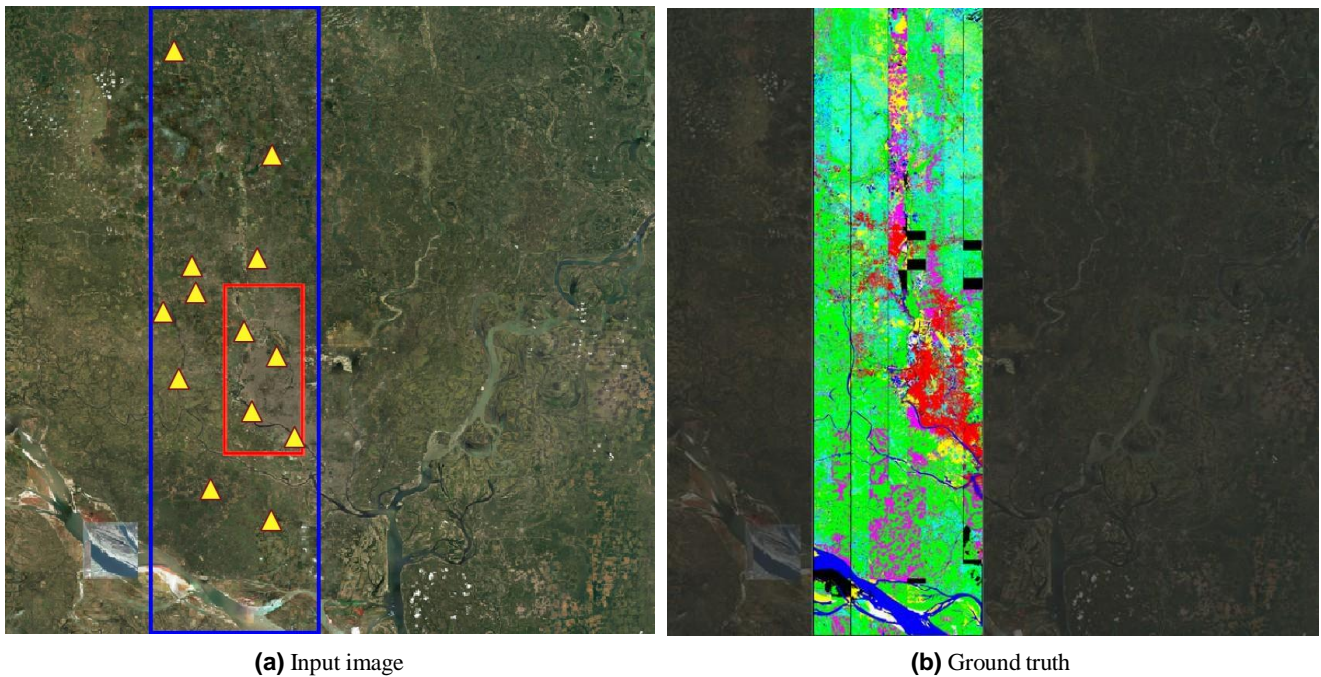

**(a)** Input image  **(b)** Ground truth

**Figure 2.** Full Bing image and the ground truth of Dhaka region.

Thus, each sub-image will be of size 1500 X 1500. Then, the sub-images from the same grid column are put in a particular folder. Therefore, each folder contains 33 images. This work focuses only on Dhaka city and its surrounding area. So, among these 33 folders, we have annotated folders from number 8 to number 19, That is, 12 folders, each containing 33 images. These 12 folders contain 1500x1500 images from the blue rectangle. For each folder, two people were assigned to do the annotations. So, a total of 24 people were given for this annotation task.

### 3.3.1 Annotation with eCognition
After loading the rule set (the rule-set is defined by the annotators for eleven classes) and a particular sub-image into eCognition software, the annotation process is started by first segmenting the image into small polygons. The segmentation is done using the multi-resolution segmentation algorithm and using the image layer weights 1.1.1. For the composition of the homogeneity criterion, the shape was set to 0.1, while compactness was set to 0.5. These polygons are interactive, and by selecting these polygons, we can set the color according to that class. The whole process is done using the eCognition software. Please See figure 3 where the sub-image and the rules are loaded in eCognition. Figure 4 shows a segmented sub-image using the eCognition.

After labeling all the polygons of the entire sub-image, a merge function is used for all the classes to merge the adjacent polygons of the same classes. Lastly, the export function is executed from the rule-set to export the final image. The final output of the ground truth is given in figure 5.

## 3.4 Annotation Training
Multiple training sessions have been conducted to teach the entire annotation process to the annotators. Training sessions have to parts: 1) technical session, 2) non-technical session.

### 3.4.1 Technical Session
In the technical session, the authors showed the technical details and the procedures to perform the annotations. In this session, the authors show the use of eCognition, the creation of rule sets, the segmentation, exporting the final output etc. The authors have made a tutorial video so that the annotators can watch it anytime during their annotation duration. The link for the tutorial video has been provided in the footnote 1.

### 3.4.2 Non-technical Sessions
Multiple discussion sessions have been conducted by the authors and the expert to help annotators identify different LULC areas. During each session, the experts and authors show a significant number of examples of above mentioned LULC classes.

---
[1]Link for annotation tutorial video: https://www.youtube.com/watch?v=psQwvRDxuTo

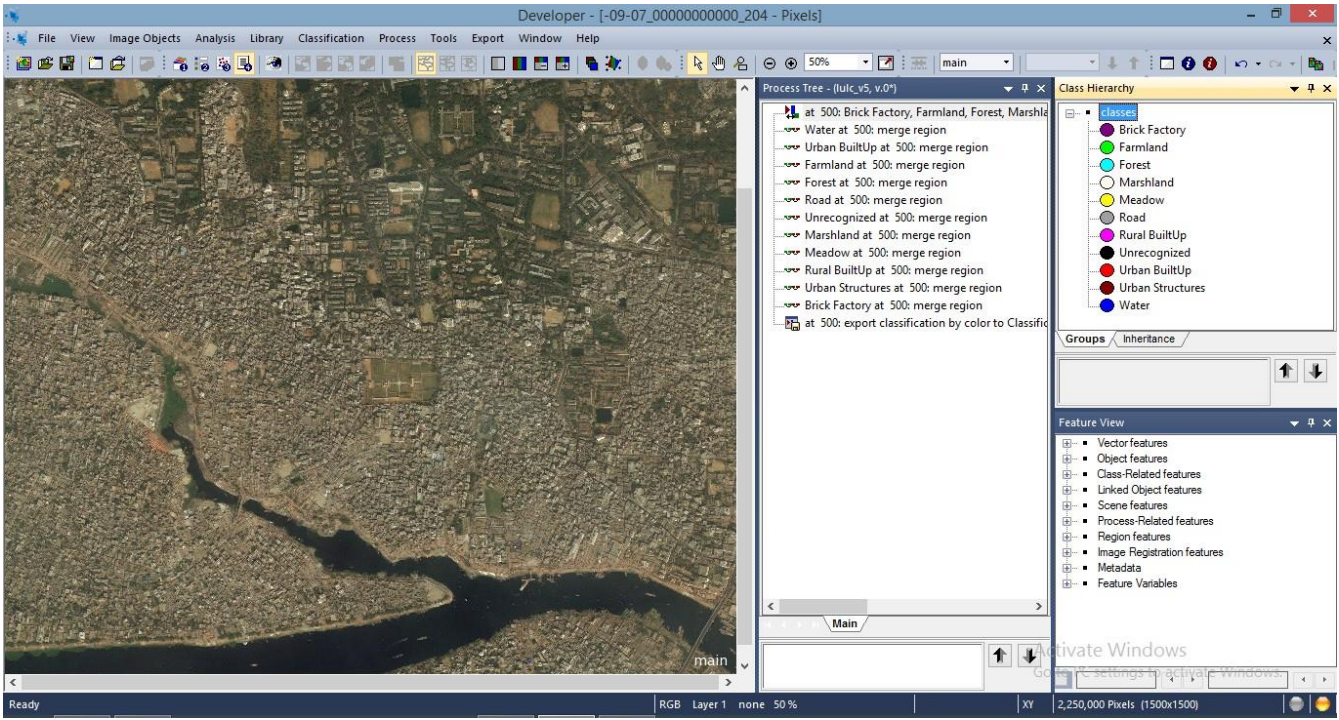

**Figure 3.** Original Image Block Upload.

They also have clarified the key differences among various LULC classes. The expert and the authors also answered questions from annotators.

### 3.5 Annotation Validation and Data Reliability
To ensure the reliability of the annotation, we have followed three steps described below.

#### 3.5.1 Initial annotation process

**Table 4.** Stage-1: Group-1 vs Group-2 annotation agreement/disagreement in percentage

| | | Group 2 | | | | | | | | | |
|---|---|---|---|---|---|---|---|---|---|---|---|
| | | Unreco-gnized | Water | Farm-land | Forest | Urban Structure | Brick Factory | Road | Urban Builtup | Rural Builtup | Marsh-land | Mea-dow |
| Group 1 | Unreco-gnized | 98.19 | 0.2 | 0.07 | 0.2 | 0.1 | 0 | 0 | 0.14 | 0.8 | 0.2 | 0.1 |
| | Water | 0.35 | 89.64 | 3.4 | 0.82 | 0 | 0 | 0 | 0 | 0 | 5.66 | 0.13 |
| | Farmland | 0.28 | 0.22 | 93.49 | 1.3 | 0.2 | 0 | 0.86 | 0 | 0.25 | 1.2 | 2.2 |
| | Forest | 0.02 | 0.01 | 0.3 | 95.24 | 0 | 0 | 0 | 0 | 3.93 | 0.1 | 0.4 |
| | Urban Structure | 0.15 | 0 | 0.73 | 0.12 | 92.97 | 0 | 2.3 | 3.73 | 0 | 0 | 0 |
| | Brick Factory | 0.02 | 0 | 0 | 0 | 2.22 | 95 | 1.9 | 0 | 0 | 0 | 0.86 |
| | Road | 1 | 0 | 0 | 0 | 3.78 | 0 | 88.18 | 0.12 | 0 | 0 | 6.92 |
| | Urban Builtup | 0.54 | 0 | 0 | 0 | 2.31 | 0.32 | 2.74 | 94.08 | 0.01 | 0 | 0 |
| | Rural Builtup | 0 | 0.08 | 1.5 | 2.5 | 0 | 0 | 0 | 1.1 | 94.82 | 0 | 0 |
| | Marshland | 0.23 | 4.73 | 0.12 | 0.08 | 0 | 0 | 0 | 0 | 0.05 | 93.9 | 0.89 |
| | Meadow | 0 | 0.18 | 0.32 | 0 | 0 | 0 | 4.8 | 0 | 3.33 | 1.5 | 89.87 |

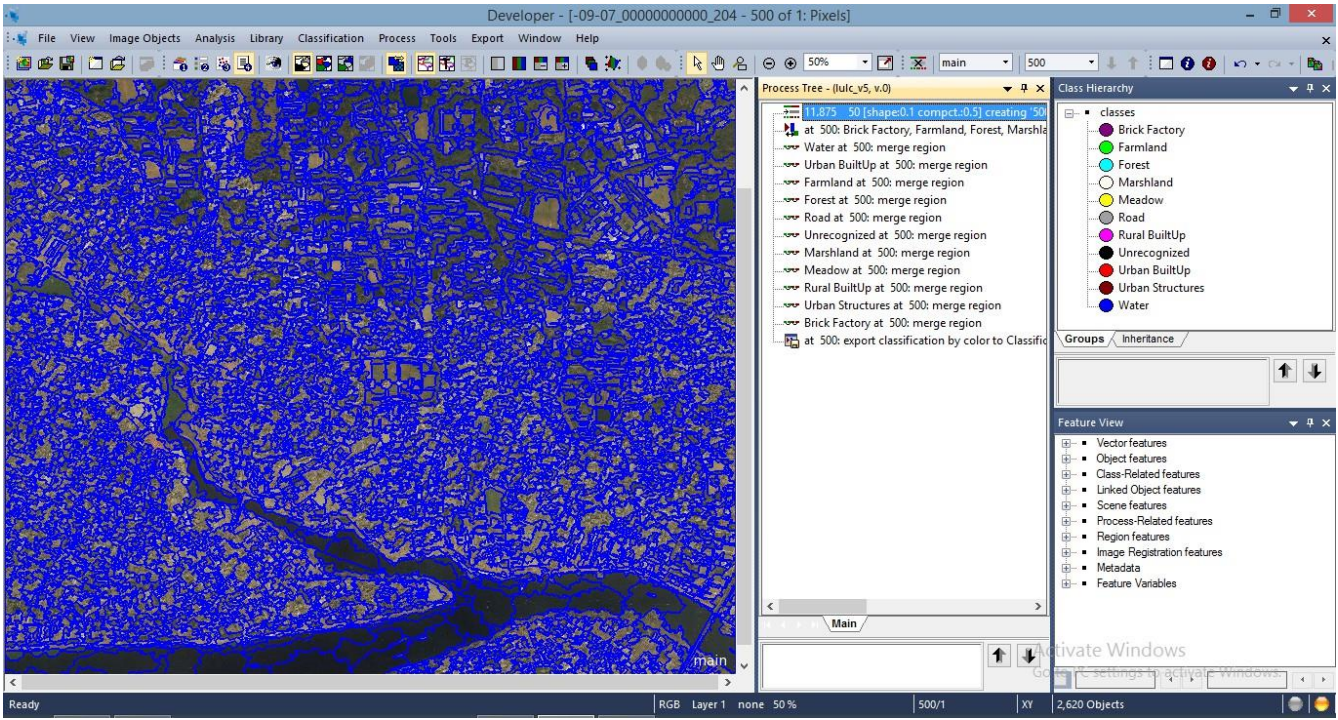

**Figure 4.** Segmenting image into small polygons

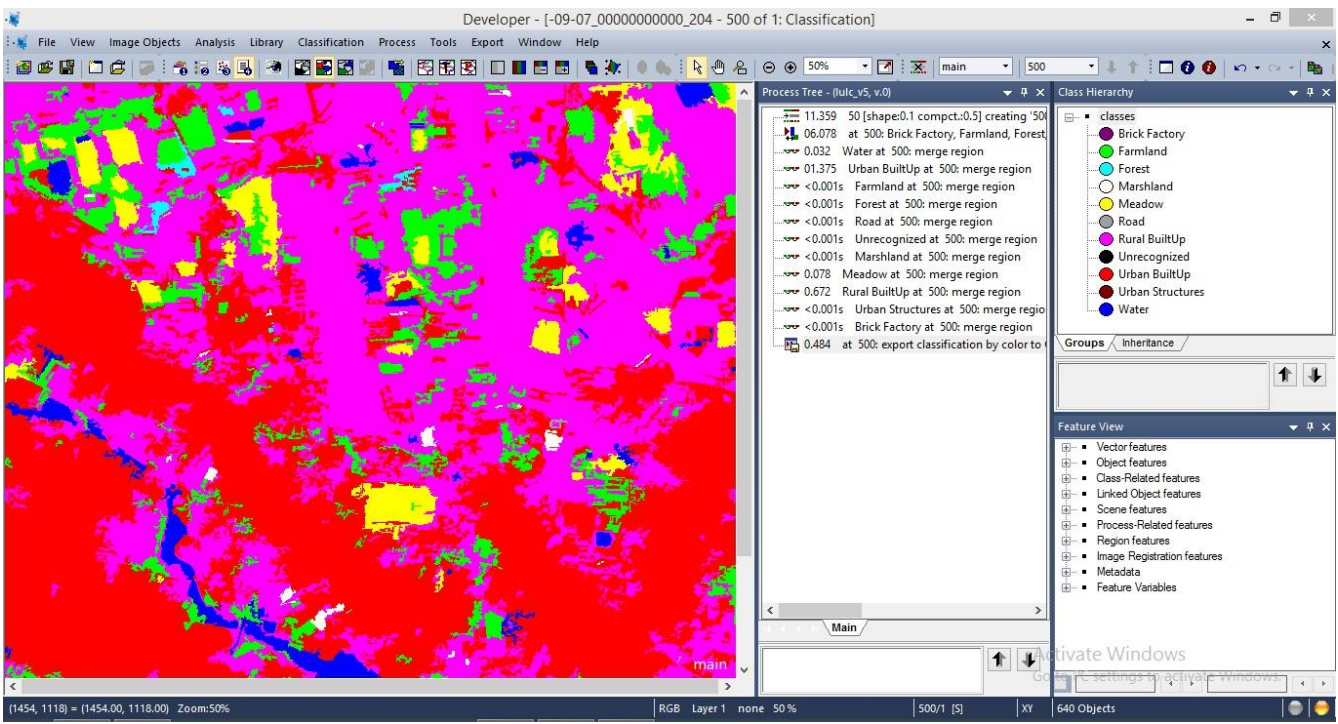

**Figure 5.** Exporting after fully labeling the ground truth

**Table 5.** Stage-1: Group-1 vs Group-2 agreement/disagreement in number of pixels

|  |  | Group 2 | | | | | | | | | | |
|---|---|---|---|---|---|---|---|---|---|---|---|---|
|  |  | Unreco-gnized | Water | Farmland | Forest | Urban Structure | Brick Factory | Road | Urban Builtup | Rural Builtup | Marsh land | Meadow |
| Group 1 | Unreco-gnized | 25534790 | 52011 | 18204 | 52011 | 26005 | 0 | 0 | 36408 | 208044 | 52011 | 26005 |
|  | Water | 140891 | 36084147 | 1368654 | 330087 | 0 | 0 | 0 | 0 | 0 | 2278406 | 52331 |
|  | Farm land | 757313 | 595032 | 252861346 | 3516095 | 540938 | 0 | 2326032 | 0 | 676172 | 3245626 | 5950315 |
|  | Forest | 29704 | 14852 | 445563 | 141451261 | 0 | 0 | 0 | 0 | 5836870 | 148521 | 594083 |
|  | Urban Structure | 6195 | 0 | 30148 | 4956 | 3839583 | 0 | 94988 | 154046 | 0 | 0 | 0 |
|  | Brick Factory | 317 | 0 | 0 | 0 | 35163 | 1504716 | 30094 | 0 | 0 | 0 | 13622 |
|  | Road | 40776 | 0 | 0 | 0 | 154132 | 0 | 3595586 | 4893 | 0 | 0 | 282167 |
|  | Urban Builtup | 262714 | 0 | 0 | 0 | 1123832 | 155682 | 1333030 | 45770593 | 4865 | 0 | 0 |
|  | Rural Builtup | 0 | 55755 | 1045400 | 1742333 | 0 | 0 | 0 | 766627 | 66083213 | 0 | 0 |
|  | Marsh land | 21858 | 449515 | 11404 | 7603 | 0 | 0 | 0 | 0 | 4752 | 8923785 | 84581 |
|  | Meadow | 0 | 118100 | 209956 | 0 | 0 | 0 | 3149345 | 0 | 2184858 | 984170 | 58964921 |

In stage 1, two groups of annotators annotated the entire dataset manually at the pixel level. The two groups of annotators annotated the same area to make the labels accurate. In this stage, two groups of annotators listed in table **??** annotate all of the images separately. Then, we calculated the pixel-wise agreement matrix for annotator group 1 vs. annotator group 2. Table 4 shows the percentage of pixel-wise agreement/disagreement after the stage 1 annotation process. In the table 5, we provide an agreement/agreement table with the number of pixels after the stage 1 annotation process.

In stage 1, the unrecognized part has been the most agreed portion of this label, which stands at 98.19 percent. However, there are disagreements across all the classes except the Brick and Road class. For the Water class, the disagreement is a lot for Farmland and Marshland, with 3.4% and 5.66%, respectively. The reason for this disagreement is a vast portion of farms in Bangladesh are beside the riverside for farming benefits. So in many areas, the water regions get merged with farmland areas. This causes an area of confusion and overlaps between the two classes. Moreover, most ponds and lakes in Bangladesh are full of weeds; the greenery floating on the water can be confused with farmland by the naked eye. This is why annotators from the two groups had disagreements in this area. Additionally, some water regions in Bangladesh may be big rivers such as Padma, Jamuna, etc., or small rivers; many carry murky waters that look the same color as the soil in satellite imagery in many regions. This is why it can be confusing to deduce a land area, whether it is marshland or water. Because of these confusions, the water region had an agreement of 89.64 percent. For the marshland class, the marshland class is causing 4.73 percent disagreement among the annotators. For the farmland class, the disagreed areas are mainly with the class: roads, rural built-up marshland, and meadow. Since many roads are beside and inside farms, it becomes difficult to differentiate the overlapping areas. The same reason goes for the other classes as well mentioned before. Many rural built-ups are beside farms, so the same overlapping issues occur. As mentioned, there are rivers beside farms, so naturally, there are marshlands around those areas. Because of flood in the rainy seasons, these marshlands increases due to the overflow of river water. So farms and marshlands get mixed up, creating an area of confusion. Also, when the crops are young or harvested, confusion occurs between the farmland and meadow classes. During that brief period, the farmlands can be mistaken for meadow areas. As for the forest region, it has a minor confusion area except in rural built-ups. In Bangladesh, houses in rural areas are often surrounded by trees. Because of this, many overlapping regions between these two classes occur as part of the trees are present in the surrounding area of the built-ups. That's why these regions cause a disagreement of 3.93 percent. In our annotation, different categories of man-made structures are included. So far, we discussed the disagreement with rural built-up. On the other hand, urban structures, brick factories, roads, and urban built-up can also be confused. For example, the brick factory areas contain the brick factory itself and roads too, so it becomes difficult to differentiate them in many cases. Additionally, some roads are closely situated with urban structures or built-ups, and these can be difficult to differentiate in many areas. The polygons created for this annotation task often overlap in two classes, especially for these man-made structures. So it becomes difficult for the annotators to differentiate both classes accurately; hence, disagreements in these areas are increased. A large portion of disagreement is caused by road and rural built-up. Since most meadow areas are in rural areas and roads in rural Bangladesh are mostly cleaned soil paths, it becomes hard to distinguish between soil-based roads with meadow areas. So in stage 1, the agreement was lowest for Road at 88.18 percent and forest being highest with 95.24 percent. Examples of disagreement issues are depicted and explained in figure 6 and 7. Many of the disagreements are resolved in the second stage, and finally, some disagreements are correctly resolved by our experts in the final stage.

### 3.5.2 Intermediate Annotation Process

**Table 6.** Stage-2: Group-1 vs Group-2 annotation agreement/disagreement in percentage

|  |  | Group 2 | | | | | | | | | | |
|---|---|---|---|---|---|---|---|---|---|---|---|---|
|  |  | Unreco-gnized | Water | Farm-land | Forest | Urban Structure | Brick Factory | Road | Urban Builtup | Rural Builtup | Marsh-land | Meadow |
| Group 1 | Unreco-gnized | 99.06 | 0.2 | 0.02 | 0.12 | 0.04 | 0 | 0 | 0.06 | 0.2 | 0.2 | 0.1 |
|  | Water | 0.35 | 97.37 | 0.09 | 0.04 | 0 | 0 | 0 | 0 | 0 | 2.02 | 0.13 |
|  | Farmland | 0.28 | 0.22 | 96.58 | 1.1 | 0.2 | 0 | 0.86 | 0 | 0.25 | 0.42 | 0.09 |
|  | Forest | 0.02 | 0.01 | 0.02 | 98.72 | 0 | 0 | 0 | 0 | 1.08 | 0.1 | 0.05 |
|  | Urban Structure | 0.15 | 0 | 0.31 | 0.12 | 95.9 | 0 | 2.69 | 0.83 | 0 | 0 | 0 |
|  | Brick Factory | 0.02 | 0 | 0 | 0 | 0.19 | 97.03 | 1.9 | 0 | 0 | 0 | 0.86 |
|  | Road | 0.04 | 0 | 0 | 0 | 1.78 | 0 | 95.46 | 0.12 | 0 | 0 | 2.6 |
|  | Urban Builtup | 0.33 | 0 | 0 | 0 | 1.21 | 0.32 | 1.28 | 96.85 | 0.01 | 0 | 0 |
|  | Rural Builtup | 0 | 0.08 | 1.3 | 1.1 | 0 | 0 | 0 | 1.1 | 96.42 | 0 | 0 |
|  | Marshland | 0.23 | 2.68 | 0.11 | 0.08 | 0 | 0 | 0 | 0 | 0.05 | 96.61 | 0.24 |
|  | Meadow | 0 | 0.18 | 0.32 | 0 | 0 | 0 | 2.2 | 0 | 2.01 | 0.09 | 95.2 |

**Table 7.** Stage-2: Group-1 vs Group-2 annotation agreement/disagreement in number of pixels

|  |  | Group 2 | | | | | | | | | | |
|---|---|---|---|---|---|---|---|---|---|---|---|---|
|  |  | Unreco-gnized | Water | Farmland | Forest | Urban Structure | Brick Factory | Road | Urban Builtup | Rural Builtup | Marsh-land | Meadow |
| Group 1 | Unreco-gnized | 25761037 | 52011 | 5201 | 31207 | 10402 | 0 | 0 | 15603 | 52011 | 52011 | 26005 |
|  | Water | 140891 | 39195821 | 36229 | 16102 | 0 | 0 | 0 | 0 | 0 | 813141 | 52331 |
|  | Farmland | 757313 | 595032 | 261218834 | 2975158 | 540938 | 0 | 2326032 | 0 | 676172 | 1135969 | 243422 |
|  | Forest | 29704 | 14852 | 29704 | 146619787 | 0 | 0 | 0 | 0 | 1604025 | 148521 | 74260 |
|  | Urban Structure | 6195 | 0 | 12803 | 4956 | 3960589 | 0 | 111095 | 34278 | 0 | 0 | 0 |
|  | Brick Factory | 317 | 0 | 0 | 0 | 3009 | 1536870 | 30094 | 0 | 0 | 0 | 13622 |
|  | Road | 1631 | 0 | 0 | 0 | 72580 | 0 | 3892432 | 4893 | 0 | 0 | 106016 |
|  | Urban Builtup | 160547 | 0 | 0 | 0 | 588674 | 155682 | 622729 | 47118217 | 4865 | 0 | 0 |
|  | Rural Builtup | 0 | 55755 | 906013 | 766627 | 0 | 0 | 0 | 766627 | 67198306 | 0 | 0 |
|  | Marshland | 21858 | 254694 | 10454 | 7603 | 0 | 0 | 0 | 0 | 4752 | 9181329 | 22808 |
|  | Meadow | 0 | 118100 | 209956 | 0 | 0 | 0 | 1443450 | 0 | 1318788 | 59050 | 62462006 |

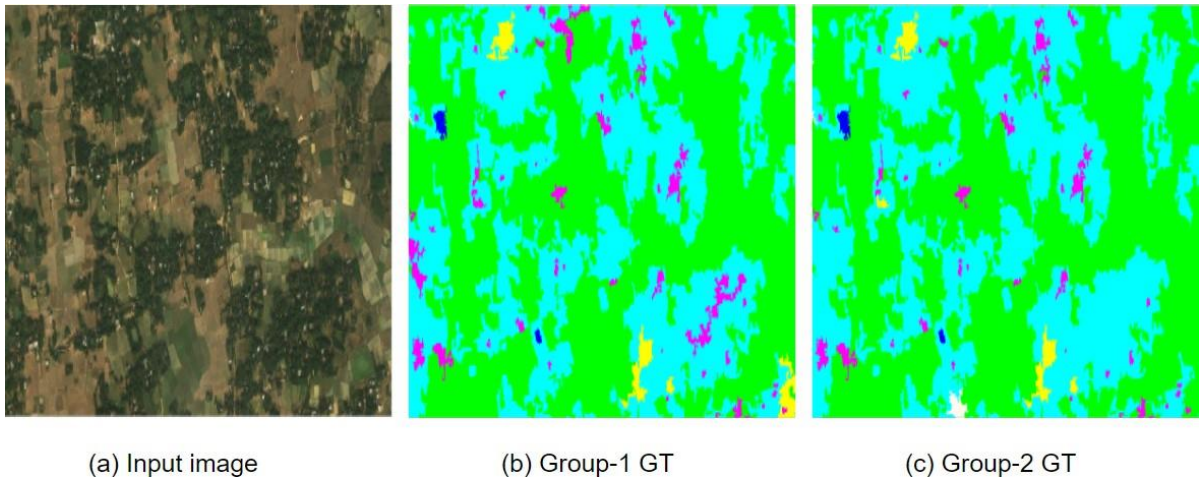

**Figure 6.** Disagreement in annotation between Group 1 and Group 2 members. Some of the water areas have been missed by Group 2

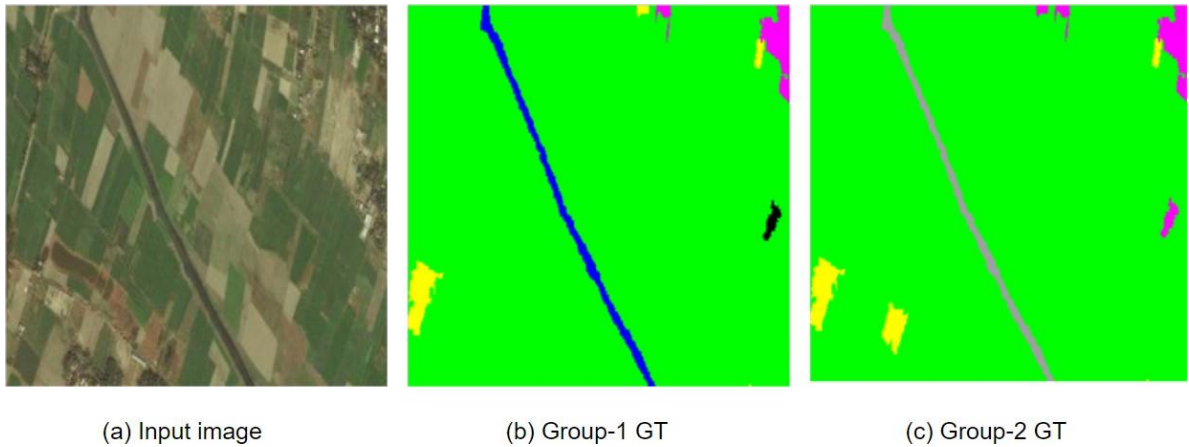

**Figure 7.** Disagreement in annotation between Group 1 and Group 2 members. One of the group 1 member annotated the long thin region as water (blue), whereas one group 2 member marked it as road. In Bangladesh, there are a lot of places where roads or rivers pass through vast farmland areas.

For stage 2, the disagreement areas of the first stage were separated from the whole dataset. Attention to detail was given to these disagreement regions. As mentioned before, the reason for these disagreements was confusion about the land cover like soil color, water, roads with the same color and texture as meadow, etc., and the other reason was the overlapping of two classes. To overcome these issues, the disagreed regions were observed more with more time and attention. we identified all the regions (square with 200x200 pixels) where annotator group 1 and annotator group 2 have disagreements with the pixels. We have discussed with the corresponding two annotators that have performed annotations on that specific region. After the discussion, the two annotators performed their annotations on the same region again. Also, the initial method of using micro polygons to annotate with eCognition was replaced with the photoshop tool with a pen on an iMac (24-inch screen) to tackle the overlapping class regions at the pixel level. Table 6 shows the percentage of pixel-wise agreement/disagreement after stage 2. In the table 7, we provide an agreement/agreement table with the number of pixels after stage 2. After finishing stage 2 with an understanding of the mistakes and confusion of stage 1, the disagreement percentage was reduced significantly. The highest disagreement percentage in stage 2 was 4.2 percent. The strategy taken for stage 2 effectively reduced the disagreement percentage.

#### 3.5.3 Final Annotation
For stage 3, we take support from a couple of geographic information system (GIS) experts that have enormous experience with LULC annotations. At first, the agreed portion of the dataset was taken, i.e., the area both group-1 and group-2 agreed on. Since the agreed portion of the dataset was thought to be highly accurate due to the agreement of both groups, only 10 percent of the agreed portion of each class was taken and given to the expert group for further verification. After the annotation of the experts, the 10 percent of the area was matched along with the group-1 and group-2 agreement area, only the urban structure class was below 99 percent agreement which is 98.585 percent. All the other classes were above 99 percent agreement. Because of such a high agreement percentage, all portions of the agreed area of group 1 and group 2 were passed as accurate and did not need further accuracy verification. Table 8 shows the percentage of pixel-wise agreement between Group 1 and Group 2 decisions vs experts' opinions in stage 3. In the table 9, we provide agreement/agreement table with the number of pixels in the stage 3.

**Table 8.** Stage-3: Group-1 and 2 agreement vs. Expert annotation in percentage

| | | Expert | | | | | | | | | | |
|---|---|---|---|---|---|---|---|---|---|---|---|---|
| | | Unrecognized | Water | Farmland | Forest | Urban Structure | Brick Factory | Road | Urban Builtup | Rural Builtup | Marshland | Meadow |
| Both Group 1 and Group 2 Agree | Unrecognized | 99.585 | 0.015 | 0.02 | 0.1 | 0.01 | 0 | 0 | 0.05 | 0.1 | 0.02 | 0.1 |
| | Water | 0.01 | 99.15 | 0 | 0 | 0.02 | 0.05 | 0 | 0.15 | 0.5 | 0.1 | 0.02 |
| | Farmland | 0.01 | 0.02 | 99.46 | 0.05 | 0.1 | 0.15 | 0 | 0 | 0.04 | 0.12 | 0.05 |
| | Forest | 0.02 | 0.01 | 0.03 | 99.13 | 0.2 | 0.05 | 0.15 | 0 | 0.25 | 0.13 | 0.03 |
| | Urban Structure | 0.015 | 0 | 0.03 | 0.1 | 98.585 | 0.05 | 0.2 | 0.3 | 0.2 | 0.5 | 0.02 |
| | Brick Factory | 0 | 0.5 | 0 | 0 | 0.05 | 99.23 | 0 | 0.12 | 0.1 | 0 | 0 |
| | Road | 0.01 | 0 | 0 | 0 | 0.1 | 0 | 99.39 | 0.5 | 0 | 0 | 0 |
| | Urban Builtup | 0.04 | 0 | 0 | 0 | 0.04 | 0.03 | 0.07 | 99.81 | 0.01 | 0 | 0 |
| | Rural Builtup | 0.014 | 0.09 | 0.15 | 0.05 | 0 | 0 | 0 | 0.03 | 99.666 | 0 | 0 |
| | Marshland | 0.01 | 0.04 | 0.3 | 0.2 | 0 | 0 | 0 | 0 | 0 | 99.41 | 0.04 |
| | Meadow | 0.05 | 0.02 | 0.12 | 0.05 | 0 | 0 | 0 | 0 | 0.05 | 0.19 | 99.52 |

**Table 9.** Stage-3: Group-1 and 2 agreement vs. Expert annotation in number pixel

| Total Pixels Selected for Validation (10% of Group 1 and 2 Agree) | | | Expert | | | | | | | | | | |
|---|---|---|---|---|---|---|---|---|---|---|---|---|---|
| | | | Unrecognized | Water | Farmland | Forest | Urban Structure | Brick Factory | Road | Urban Builtup | Rural Builtup | Marshland | Meadow |
| 2576104 | Both Group 1 and Group 2 Agree | Unrecognized | 2565413 | 386 | 515 | 2576 | 258 | 0 | 0 | 1288 | 2576 | 515 | 2576 |
| 3919582 | | Water | 392 | 3886266 | 0 | 0 | 784 | 1960 | 0 | 5879 | 19598 | 3920 | 784 |
| 26121883 | | Farmland | 2612 | 5224 | 25980825 | 13061 | 26122 | 39183 | 0 | 0 | 10449 | 31346 | 13061 |
| 14661979 | | Forest | 2932 | 1466 | 4399 | 14534419 | 29324 | 7331 | 21993 | 0 | 36655 | 19061 | 4399 |
| 396059 | | Urban Structure | 59 | 0 | 119 | 396 | 390455 | 198 | 792 | 1188 | 792 | 1980 | 79 |
| 153687 | | Brick | 0 | 768 | 0 | 0 | 77 | 152504 | 0 | 184 | 154 | 0 | 0 |
| 389243 | | Road | 39 | 0 | 0 | 0 | 389 | 0 | 386869 | 1946 | 0 | 0 | 0 |
| 4711822 | | Urban Builtup | 1885 | 0 | 0 | 0 | 1885 | 1414 | 3298 | 4702869 | 471 | 0 | 0 |
| 6719831 | | Rural Builtup | 941 | 6048 | 10080 | 3360 | 0 | 0 | 0 | 2016 | 6697386 | 0 | 0 |
| 918133 | | Marshland | 92 | 367 | 2754 | 1836 | 0 | 0 | 0 | 0 | 0 | 912716 | 367 |
| 6246200 | | Meadow | 3123 | 1249 | 7495 | 3123 | 0 | 0 | 0 | 0 | 3123 | 11868 | 6216219 |

**Table 10.** Stage-3: Group-1 and 2 disagreements vs. Expert annotation in number of pixels

| Total Pixels Selected for Validation (100% of 1 and 2 Disagree) | | | Expert | | | | | | | | | | |
|---|---|---|---|---|---|---|---|---|---|---|---|---|---|
| | | | Unrecognized | Water | Farmland | Forest | Urban Structure | Brick Factory | Road | Urban Builtup | Rural Builtup | Marshland | Meadow |
| 244452 | Both Group 1 and Group 2 Disagree | Unrecognized | 243437 | 37 | 49 | 244 | 24 | 0 | 0 | 122 | 244 | 49 | 244 |
| 1058694 | | Water | 106 | 1049695 | 0 | 0 | 212 | 529 | 0 | 1588 | 5293 | 1059 | 212 |
| 9250035 | | Farmland | 925 | 1850 | 9172335 | 32375 | 9250 | 13875 | 0 | 0 | 3700 | 11100 | 4625 |
| 1901067 | | Forest | 380 | 190 | 6844 | 1878254 | 3802 | 951 | 2852 | 0 | 4753 | 2471 | 570 |
| 169327 | | Urban Structure | 254 | 0 | 847 | 169 | 165906 | 85 | 339 | 508 | 339 | 847 | 34 |
| 47042 | | Brick Factory | 0 | 235 | 0 | 0 | 24 | 46680 | 0 | 56 | 47 | 0 | 0 |
| 185121 | | Road | 19 | 0 | 0 | 0 | 185 | 0 | 183992 | 926 | 0 | 0 | 0 |
| 1532498 | | Urban Builtup | 613 | 0 | 0 | 0 | 6130 | 4597 | 10727 | 1510276 | 153 | 0 | 0 |
| 2495021 | | Rural Builtup | 3493 | 2246 | 3743 | 1248 | 0 | 0 | 0 | 749 | 2483544 | 0 | 0 |
| 322169 | | Marshland | 32 | 1289 | 967 | 644 | 0 | 0 | 0 | 0 | 0 | 319108 | 129 |
| 3149345 | | Meadow | 1575 | 1249 | 7495 | 3123 | 0 | 0 | 0 | 0 | 3123 | 11868 | 6216219 |

Finally, what is left is to annotate the disagreed portion of the dataset between group 1 and group 2. The entire portion of the disagreed area was given to the expert. The whole disagreed area was annotated by the experts from scratch. Table 10 shows the number of pixels for agreement/disagreement between group 1 - group 2 disagreement vs. expert's opinion. Since the expert annotated the entire portion, we keep the annotations from the expert's opinion for the pixels identified in table 10. After completing this portion, the whole dataset was finally finished at stage 3.

### 3.6 In Situ Visual Assessment

The accurate classification of land use and land cover (LULC) is crucial for various applications such as urban planning, environmental monitoring, and resource management. To ensure the reliability of LULC dataset annotations, an on-site validation process was conducted in the Dhaka Division. This involved physically visiting different locations within the division and comparing the dataset's annotations. A comprehensive plan was developed to cover a diverse range of landscapes and land uses within the Dhaka Division. The selection of locations was based on factors such as urban, rural, agricultural, industrial, and natural areas. Necessary equipment was gathered, including a digital camera or smartphone with high spatial resolution capabilities, a navigation app Gaia GPS was used to mark the location visited, notebooks for recording observations, and any other required accessories. On-site visits were carried out, involving visits to various places identified during the planning stage. Upon reaching a designated point, the specific land use and land cover characteristics were carefully observed and noted down. This included details such as urban infrastructure, vegetation, water bodies, agricultural fields, and any other distinctive features. High-resolution photographs were taken to capture the actual state of the land at that moment. These photographs aimed to visually document the ground conditions and serve as a reference for comparison with the dataset's annotations. GPS coordinates were recorded for each location to precisely match the on-site observations with the corresponding points in the LULC dataset. This ensured accurate correlation between ground truth data and the dataset's annotations. Our validation team visited 13 diverse locations marked in figure 2 across the Dhaka Division, deliberately covering a spectrum of land use types. These sites included urban areas, rural landscapes, riverbanks, agricultural fields, industrial zones, and natural habitats. By doing so, we aimed to capture the full range of land cover features present in the region.

The collected photographs and ground observations were compared with the existing annotations in the LULC dataset. Four sample photographs are shown in the fig 8 that were taen during the onsite visits. Discrepancies between the dataset's labels and real-world observations were noted for further analysis. The observations made during the on-site visits were used to assess the accuracy of the dataset's annotations. Where discrepancies were identified, the labels have been updated and have been improved to reflect the true LULC conditions more accurately. The on-site validation process played a pivotal role in ensuring the accuracy and reliability of the LULC dataset for the Dhaka Division. By directly comparing dataset annotations with real-world conditions through photography and ground observations, potential errors and inaccuracies were identified and subsequently addressed. This process enhances the dataset's utility for various applications and contributes to more informed decision-making in urban planning, environmental management, and beyond.

### 3.7 Conversion from 11 classes to 6 classes

Careful observation reveals that the percentages of Roads, Brick Factories, and Urban structures are very small (fig. 9a). Therefore, a significant class imbalance problem is present in the data for eleven classes. Hence, we combine eleven classes into six essential classes shown in the 3rd column for performing our experiments and for establishing benchmark results. In the converted classes, Farmland, Water, and Forest are kept as it was in the original ten classes. But the Urban Structure, Rural Built-up, Urban Built-up, and Brick Factory are merged into the Built-up category. Meadow and Marshland are combined into the Meadow class. The area percentages for six classes are shown in figure 9b

### 3.8 Pre-Processing

The original image is split into patches of 513x513 resolution to feed into the network. The ground truth has many regions that are difficult for the annotators to label, so those regions were classified as unrecognized. This unrecognized is the redundant part of the ground truth that needs to be removed from the input image. This is why the unknown of the ground truth is mapped on the input image to put a black color region there, which helps our model not to confuse this unrecognized class with any actual class of the input. The annotated part selected for training was not enough, so augmentation was needed, which overlapped during the grid. Finally, the images for both information and target are fed into the DeepLab V3+.

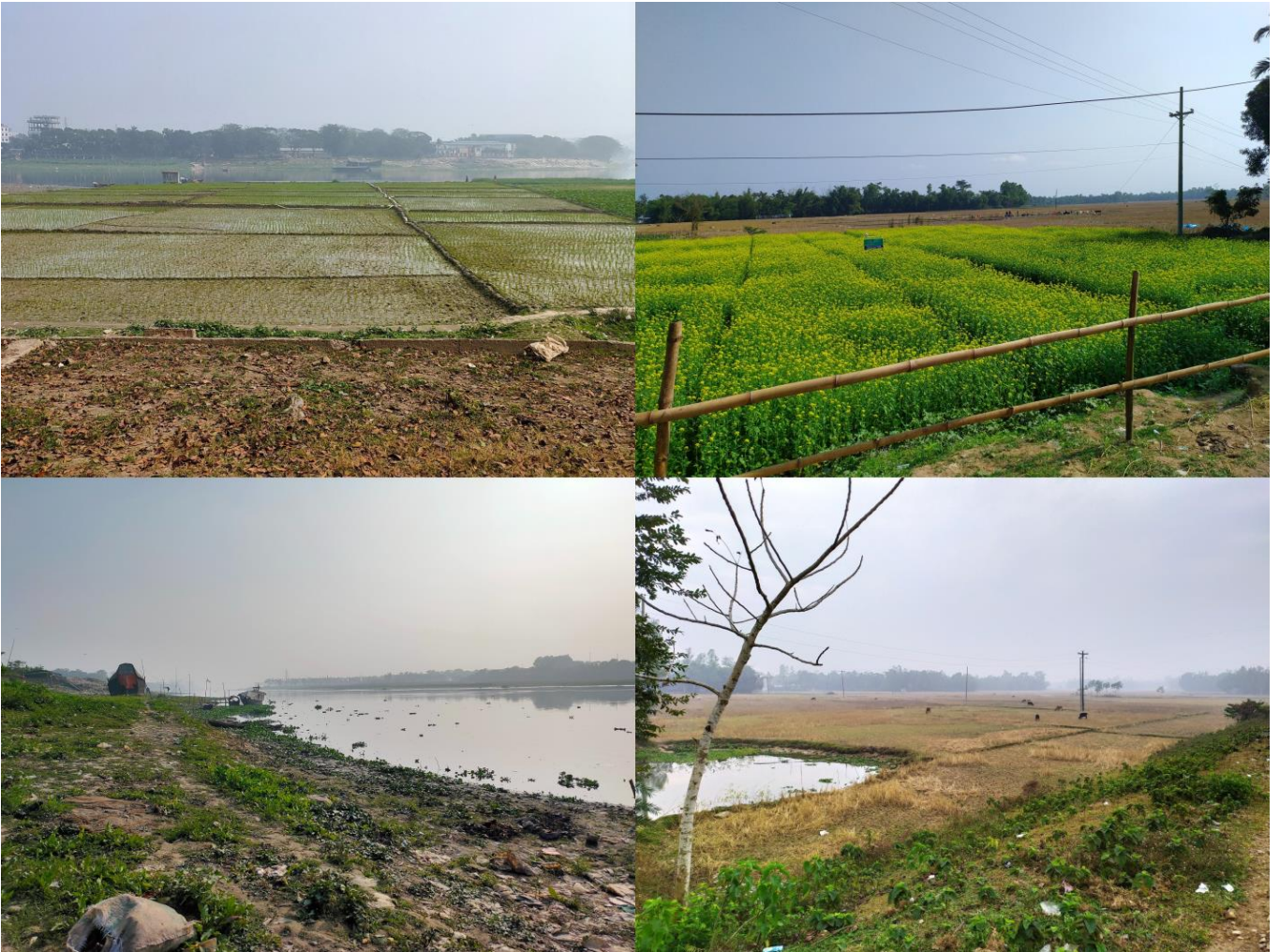

**Figure 8.** Various Regions on the Dhaka Division

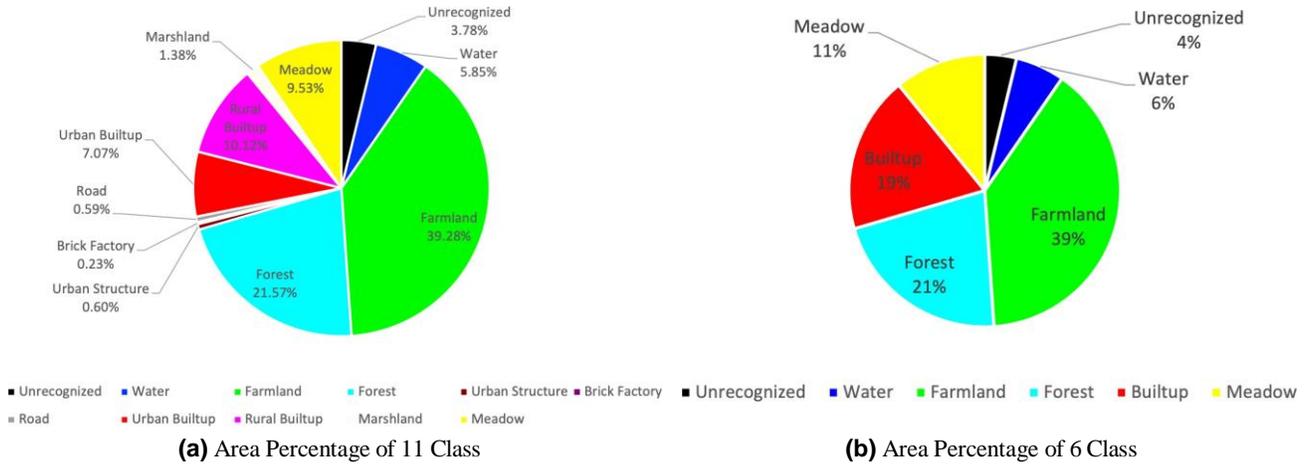

**Figure 9.** Area Percentage of the 11 class and 6 class dataset.

### 3.9 Post-Processing

After getting the output from DeepLabv3+, the images have to be reverted to three-channel data as the final softmax outputs are indexed with class labels. After decoding the images, they are stitched together to their original dimension.

## 4 Remote Sensing Data Description

Sentinel-2A satellites have 12 bands that can be combined for the desired segmentation task. As different channels capture various information about the land textures, we use different combinations from Sentinel-2A data. These combinations are formulated in a certain way to accomplish specific tasks. There are a couple of ways in which two or multiple channels are combined. The simpler one combines three different channels, such as an RGB image. The other way is to calculate the index image using a defined formula that uses two or multiple bands. Normalized vegetation index (NDVI) is a widely used example for index images. Moreover, the use of index images is essential to perform LULC. The RGB images of the index and combination images are shown in Figure 10. This dataset has been made available online here under Creative Common (CC) license: https://doi.org/10.7910/DVN/LLR3RR[61]. Please note that author names are removed to maintain the anonymity of the manuscript.

### 4.1 Index Images

Each indexed image has a specific characteristic that can detect particular classes. For example, NDVI is used to detect greenery. Vegetation in an area can be measured using this index[56]. Another index Normalized Difference Water Index(NDWI), can detect water which can be helpful for various problems. Mosquitos are a massive problem in modern times; using the NDWI water region in a city is seen where mosquitos can be born[59]. As the index image is a single channel data, we copy the calculated data into three channels to produce a three-channel input image for the deep learning model (see table 11). Since different channels are captured with various resolutions, we have used the bilinear interpolation[62] to up-sample them to 10 meters/pixel. The following index images are widely used in the LULC literature.

**Table 11.** Sentinel-2 Index and Combination RGB Channel Formation

| Image Type | Channel-1 | Channel-2 | Channel-3 |
|---|---|---|---|
| RGB | B4(Red) | B3(Green) | B2(Red) |
| ATM | B12(SWIR) | B11(SWIR) | B8(NIR) |
| FCI | B8(NIR) | B4(Red) | B3(G) |
| SWI | B12(SWIR) | B8(NIR) | B4(R) |
| NDVI | NDVI | NDVI | NDVI |
| NDWI | NDWI | NDWI | NDWI |
| RVI | RVI | RVI | RVI |

**Normalized Difference Vegetation Index(NDVI):** Because near-infrared (which vegetation strongly reflects) and red light (which vegetation absorbs), the vegetation index is suitable for quantifying the amount of greenery. High values suggest a dense canopy. The formula for the normalized difference vegetation index is:

$$NDVI = \frac{B8(NIR) - B4(R)}{B8(NIR) + B4(R)} \qquad (1)$$

**Normalized Difference Water Index(NDWI):** The moisture index is ideal for finding water stress in plants. It uses the short-wave and near-infrared to generate an index of moisture content. In general, wetter vegetation has higher values. But lower moisture index values suggest plants are under stress from insufficient moisture. The formula is:

$$NDWI = \frac{B8A(NarrowNIR) - B11(SWIR)}{B8A(NarrowNIR) + B11(SWIR)} \qquad (2)$$

**Radar Vegetation Index (RVI):** Different bands of a multi-spectral image can also be combined to bring out the vegetated areas. An example of the combination is the ratio of the NIR (near-infrared) band to the red band. It can be calculated as follows.

$$RVI = \frac{B8(NIR)}{B4(R)} \qquad (3)$$

## 4.2 Channel Combination images

Different proper channels can be combined to create combination images. For example, combining all short wave infrared bands (SWI) is used to detect minerals and moisture content[54]. Since different channels are captured with various resolutions, we have also used the bilinear interpolation[62] to up-sample them to 10 meters/pixel so that the layer stacking can be done seamlessly. From the Sentinel-2A data, four combination images were used in this paper.

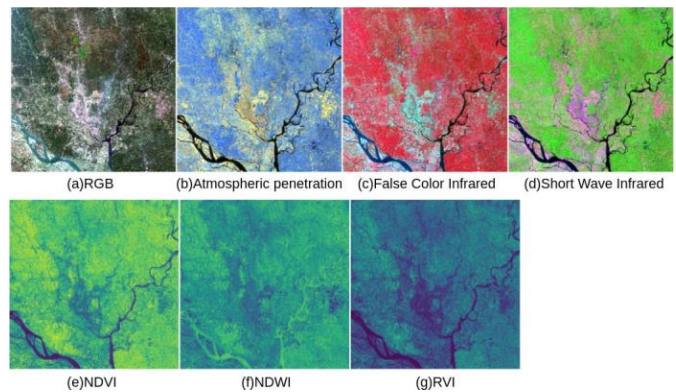

**Figure 10.** Combination and Index images generated from Sentinel-2 multispectral bands

**False Color Infrared(FCI):** The false color infrared band combination is meant to emphasize healthy and unhealthy vegetation. It combines NIR(B8), red(B4), and green (B3). The near-infrared (B8) band is especially good at reflecting chlorophyll content found in green plants. This is why denser vegetation is red and urban areas are white in a false-color infrared image.

**Atmospheric Penetration(ATM):** This band combination shows results similar to that of traditional false-color infrared photography while retaining excellent clarity[63]. It involves SWIR (B12), SWIR(B11), and NIR(B8). Thus it has no visible bands while penetrating atmospheric particles, smoke, and haze, reducing the atmospheric influence in the image. Water attenuates the NIR and the short-wave infrared (SWIR) wavelengths, ice, and snow, giving well-defined shores, coastlines, and highlighted water sources within the image.

**Short Wave Infrared(SWI):** The short-wave infrared band combination uses SWIR (B12), NIR (B8A), and red (B4). This composite shows vegetation in various shades of green. In general, darker shades of green indicate denser vegetation. But brown is indicative of bare soil and built-up areas.

**SentinelRGB:** True color composite uses visible light bands red (B04), green (B03), and blue (B02) in the corresponding red, green and blue color channels from Sentinel 2A data, resulting in a natural colored result. This is a good representation of the earth's surface as humans would see it naturally.

### 4.3 Bing RGB Image

The Bing image used for the experiment is taken at 17 zoom levels, 2.22 meters per pixel. The reason for using this image is to make a quality ground truth with details and to segment the LULC classes as correctly as possible. Because of its high resolution, the data is huge. The only difficulty working with data is the processing and training time. But it comes with the394 advantage of the detailed image, which a human can see with a naked eye to annotate correctly, and a deep learning model to segment LULC classes.

## 5 Experiments with the Deep Learning Model

In this section, we provide the technical details for model training, model testing, performance evaluation, and experimental setup. Then we describe the benchmark results that we establish in this paper. Deep Lab V3+ has been used for the LULC segmentation for all the experiments.

### 5.1 Train and Test Splits
The Dhaka city region (marked by a red rectangle in figure 2a) is used as a testing part, while the area outside of Dhaka city encompassed by the blue rectangle is used as training data.

### 5.2 Evaluation Metrics
We have calculated Intersect over Union(**IoU**) and **F1** scores from the confusion matrices constructed from the model's prediction and ground truths over test image pixels. We didn't use accuracy there as accuracy might be biased with the imbalanced samples of LULC classes.

We start our experiments with index images and channel-combined images. Finally, an investigation with a high-resolution image from Bing is conducted to see if the deep learning model can output more accurate detail in LULC segmentation. In this section, we describe the experimental settings.

### 5.3 LULC with Index Images
The index image is a single-channel image; we copy the single-channel information to three channels to produce a three-channel image since the deep learning architecture takes 3-channel data. In this way, we create 3-channel input images with NDVI. In the same way, 3-channel images NDWI and RVI are created. The table 11 shows how a particular index image is used to create the 3-channel input image. The training images are generated by sliding the 513 X 513 window vertically and horizontally with 70% overlapping. The batch size is eight and epoch 25; the data is trained using the DeepLabV3+. The trained model is used to test Dhaka city to get the result. For the testing, sliding windows are used with no overlapping.

### 5.4 LULC with Channel combination images
Using QGIS, three particular bands are combined to create the corresponding channel combined images. The training images are generated by sliding the 513 X 513 window vertically and horizontally with 70% overlapping. The batch size is eight and epoch 25; the data is trained using the DeepLabV3+. No overlapping is used during the testing.

### 5.5 LULC with Full Bing RGB
The full Bing image has been used here. The training images are generated by sliding the 513 X 513 window vertically and horizontally with 70% overlapping. The batch size is eight and epoch 25; the data is trained using the DeepLabV3+.

## 6 Results
In this section, we describe all the experimental results.

### 6.1 Index images
Among the three index images, NDWI has the highest IoU (0.20) and F1 score (0.32) for the forest. NDWI has the highest performance for water(IoU: 0.24, F1 score: 0.51). In contrast, NDVI is winning for built-up and farmland. RVI has a little bit better F1 score for meadow. On average, NDVI does better segmentation for the five classes (average IoU: 0.30, average F1 score: 0.45).It can be easily observed that built-up and farmland are also segmented with high precision with the index images.

In the case of the Farmland class, NDVI and RVI scores were the same at 0.79 but NDVI scored higher in IoU than the other two index images with 0.35. NDWI is an index to detect water or moisture content, so our experiment was a good way to compare if it can outscore the other index images. NDWI scored higher in IoU with 0.24. So, it seems NDWI does better in detecting water regions. As for the Built-up, it is detected better in NDVI (IoU: 0.52, and F1: 0.68) and Meadow in RVI (IoU: 0.52, and F1: 0.68). The average score of RVI and NDVI are the same but NDVI has a higher IoU average of 0.308.

**Table 12.** Scores for 5 classes of all the image types

|  | Metric | Forest | Built-Up | Water | Farm-land | Mea-dow | Avg |
|---|---|---|---|---|---|---|---|
| NDVI | IoU | 0.17 | **0.52** | **0.68** | 0.35 | **0.51** | 0.30 |
|  | F1 | 0.17 | **0.68** | 0.27 | **0.51** | 0.50 | **0.45** |
| NDWI | IoU | **0.20** | 0.38 | 0.24 | 0.34 | 0.28 | 0.28 |
|  | F1 | **0.32** | 0.55 | 0.38 | 0.51 | 0.50 | 0.43 |
| RVI | IoU | 0.13 | 0.50 | 0.17 | 0.31 | 0.36 | 0.29 |
|  | F1 | 0.22 | 0.66 | 0.29 | 0.47 | **0.52** | 0.43 |
| ATM | IoU | 0.17 | **0.56** | 0.29 | **0.48** | 0.19 | 0.33 |
|  | F1 | 0.29 | **0.72** | 0.45 | **0.64** | 0.31 | 0.48 |
| FCI | IoU | 0.20 | 0.53 | 0.30 | 0.46 | 0.17 | 0.33 |
|  | F1 | 0.33 | 0.69 | 0.45 | 0.62 | 0.29 | 0.47 |
| SWI | IoU | **0.22** | 0.56 | **0.32** | 0.46 | 0.22 | **0.35** |
|  | F1 | **0.35** | 0.71 | **0.48** | 0.62 | 0.37 | **0.50** |
| SentinelRGB | IoU | 0.18 | 0.54 | 0.30 | 0.46 | **0.25** | 0.34 |
|  | F1 | 0.30 | 0.70 | 0.46 | 0.63 | **0.40** | 0.49 |
| BingRGB | IoU | 0.33 | 0.58 | 0.48 | 0.57 | 0.26 | 0.44 |
|  | F1 | 0.49 | 0.73 | 0.65 | 0.72 | 0.41 | 0.60 |

### 6.2 Channel combination images

Now, we describe the results for the combined images for sentinel 2A from the table 12. For the class Forest, SWI scores better than ATM in IoU with 0.22. In Built-up ATM and SWI scores even in IoU 0.56 but the accuracy is highest in SWI. Water region is better detected in SWI for IoU and ATM IoU is highest for class farmland. Finally, for the class Meadow, SWI got the highest IoU of 0.22. As for the average overall class scores, SWI has the higher average IoU score.

From the table 12, it can be easily observed that SWI has the highest average IoU (0.35) and F1 score (0.50) compared to other compositions.

The average performance of visual sentinelRGB has a little bit lower performance compared to SWI. Now, let's analyze the class-wise performance across the composition images. Built-ups and farmlands are segmented more accurately with all of the composition images. Among them, ATM achieves the best IoU:0.56 and F1 score of 0.72 while segmenting built-up. For Farmland, ATM performs the IoU: 0.48 and F1 score: 0.64. Forest and Water are segmented more correctly with SWI than other combinations. And for Meadow, the IoU: 0.25 and F1: 0.40 are achieved for the Sentinel RGB experiment. We also conclude that SWI is the combination that achieves the best average IoU and average F1 score among four combinations and three index images derived from Sentinel 2A data.

SWI has the highest score in Forest among all the sentinel-2 images created with 0.22 IoU. For Built-up ATM and SWI IoU scores are the same. SWI is also the highest scorer in the Water class for IoU. For Farmland, IoU is the highest for ATM, and for Meadow, it is RVI. The combination images seemed to score highest in all classes except the meadow class. A more detailed comparison also reveals that combination images are better than index images due to the more data and information in three different channels of combination images. Moreover, though SentinelRGB image is widely used for LULC, it cannot outperform most other combination images. But it is not the same for the index images, as it is one channel and contains less information. sentinelRGB could not outperform other combination images.

### 6.3 Bing RGB image

Here, we describe the results from BingRGB where the experiment includes the training and testing using the full RGB image captured by the Bing satellite. It can be easily noticed that the model with a full Bing RGB image outperforms the rest by significant margins. The average IoU is 0.44, and the average F1 score is 0.60. Not only with averages, but the model with

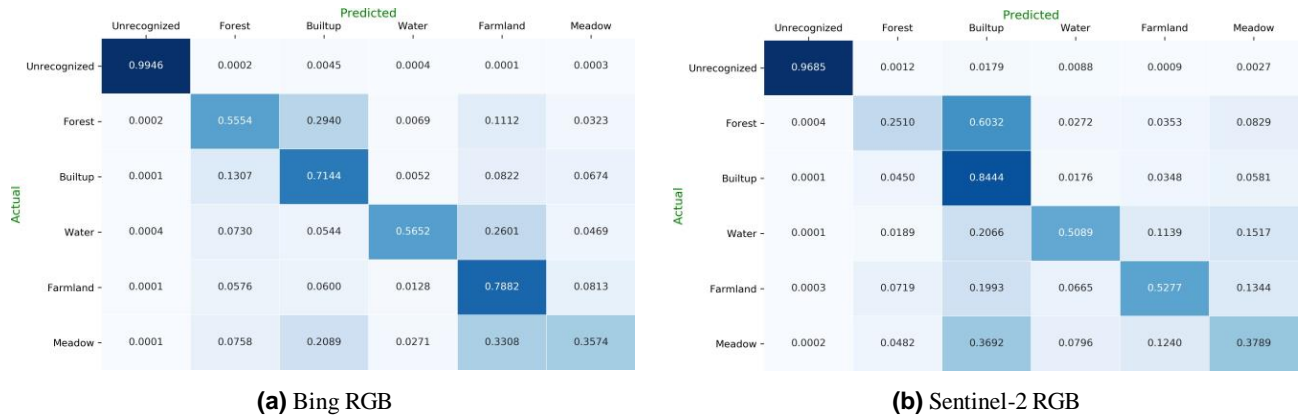

**(a)** Bing RGB

**(b)** Sentinel-2 RGB

**Figure 11.** Normalized confusion matrix of Bing RGB and Sentinel-2 RGB images.

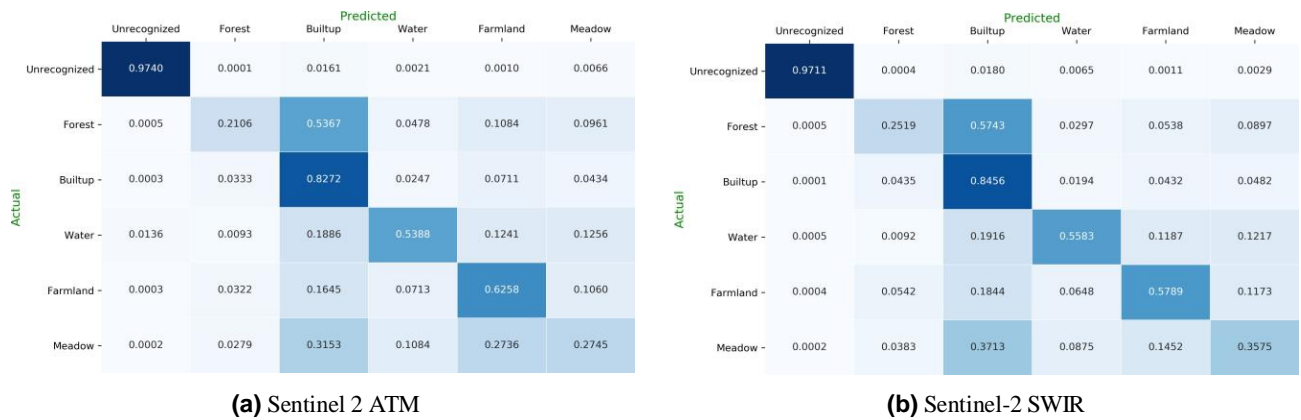

**(a)** Sentinel 2 ATM

**(b)** Sentinel-2 SWIR

**Figure 12.** Normalized confusion matrix of Sentinel-2 ATM and SWIR combination images.

BingRGB can also segment all of the five LULC classes more accurately compared to the other models.

### 6.4 Analysis of Normalized Confusion Matrix

In this subsection, we discuss the normalized confusion matrix for the combination images and BingRGB image type.

The normalized confusion matrix in Figure 11a shows the performance of DeeplabV3+ on the BingRGB dataset. Here we can see that 55% forest, 71% built-up area, 57% water, 78% farmland, and 38% meadow are correctly classified. Major inter-class confusions occurred between the classes of forest and built up which could be attributed to the proximity of vegetation and built-up in urban areas. Water is also predicted as farmland in 26% of the cases, which could be due to the color of waterbodies in urban areas where both are often similar green or brown colors instead of distinct blue water. Since meadow is an under-represented class in this dataset, it has been confused with farmland and built up more than 50% of the cases. All the classes except meadow have been classified correctly in more than 55% cases.

The normalized confusion matrix in Figure 11b shows the performance of DeeplabV3+ on the SentinelRGB dataset. Here we can see that 25% forest, 84% built-up area, 51% water, 53% farmland, and 38% meadow are correctly classified. Major inter-class confusions occurred between the classes forest and built-up where more than 60% of the forest has been classified as built-up, which could be attributed to the proximity of vegetation and built-up in urban areas as well as the inability to correctly distinguish both classes in lower resolution image where they are mushed together. Builtup in particular, did very well with around 85% accuracy. Performance in regards to water is close to the BingRGB. However, farmland did poorly where it was classified as Builtup and Meadow around 34% of the time. The performance of the meadow is almost similar to that in the BingRGB dataset. We can attribute the poor performance in most classes to the dataset's lower resolution.

Now we describe the model's performance on the Sentinel-2 ATM image. The normalized confusion matrix has been provided in Figure 12a. Farmland and water were classified more accurately in the Sentinel-2 ATM imagery dataset than in the Sentinel-2 RGB one. However, the accuracy in forest, built-up, and meadow decreased compared to RGB images. These phenomena could be attributed to the combination involving no visible bands and providing the best atmospheric penetration.

Vegetation appears in blue colors, bare soil in shades of brown and yellow. Water appears in very dark shades or black, thus well-defined. Oftentimes, the wetland is used for farming and, as a result, has better visibility in the ATM combination images.

Now we describe the model's performance on the Sentinel-2 SWI image. The normalized confusion matrix has been provided in Figure 12b. The Deeplabv3+ performance on SWI images is similar to the RGB images, with slightly better accuracy in all the classes except the meadow. Since the SWIR combination highlights the vegetation in various shades of green, in some cases, some of the meadows are classified as farmland as well. Another underlying cause could be that two out of three bands of SWIR combination are of 20m resolution. But, overall, it performed better than RGB, except with meadow.

|  | Unrecognized | Forest | Builtup | Water | Farmland | Meadow |
|---|---|---|---|---|---|---|
| Unrecognized | 0.9714 | 0.0008 | 0.0183 | 0.0071 | 0.0012 | 0.0012 |
| Forest | 0.0006 | 0.2674 | 0.5742 | 0.0283 | 0.0455 | 0.0840 |
| Builtup | 0.0002 | 0.0466 | 0.8516 | 0.0224 | 0.0430 | 0.0362 |
| Water | 0.0021 | 0.0078 | 0.2356 | 0.5616 | 0.0867 | 0.1062 |
| Farmland | 0.0005 | 0.0593 | 0.2193 | 0.0841 | 0.5243 | 0.1126 |
| Meadow | 0.0004 | 0.0394 | 0.4924 | 0.1147 | 0.1138 | 0.2394 |

**Figure 13.** Normalized confusion matrix of Sentinel-2 FCI image

Now we describe the model's performance on the Sentinel-2 FCI image. The normalized confusion matrix has been provided in Figure 13. Here we can see that FCI is classified with better accuracy for the forest, built-up, and water classes than the RGB combination. At the same time, it is almost similar in the case of farmland. However, meadow did much worse here than all the combinations discussed in this paper. Much of the meadow was classified correctly and was now classified as built-up. This could be attributed to the fact that dried-up meadow-covered land is often not as green as vegetation-covered land, and as a result, they are classified as built-up areas.

### 6.5 LULC for full Bangladesh

Since the BingRGB shows the best performance in the benchmark table, we use our BingRGB-trained DeepLabv3+ to perform LULC segmentation over all of the 64 districts of Bangladesh. The following figures 14, 15 show the outputs of automatic LULC segmentation for a couple of districts.

## 7 Detail Discussion and conclusion

In this paper, we perform several experiments to establish the benchmark results with the annotated data. The annotated data is observed sufficient to train large deep-learning models with adequate accuracy. Data from different channels from Sentinel 2A are used with the same ground truth we have generated. Experiments on combinations and index images of sentinel-2A have been done to establish the model performance on low-resolution images (10-20 meters/pixels). In the end, we use Bing satellite images as the input to the model. We have observed that LULC segmentation performance on Bing images outperforms all experiments with channel combinations and index images from Sentinel 2A. Here, we provide conclusive recommendations regarding the scope of the ground-truth annotation information with sentinel2/Bing or Google images to be used by the remote sensing community in developing countries. Deep learning models can effectively and efficiently be trained for LULC tasks with annotated high-resolution RGB images (such as Bing/Google/ESA images). However, Bing/google high resolutions images are not freely available every month for Bangladesh and other developing countries. Therefore, our annotation can use high-resolution satellite imagery (e.g., Bing/Google/ESA) when accuracy is more critical than periodic availability. For example, urban expansion or annual change detection can be monitored from Bing imagery. In contrast, crop monitoring requires periodic data availability and thus requires direct satellite imagery (e.g., Sentinel 2A) from respective vendors/providers[64]. Since high-resolution images from individual vendors/providers are not free, one can use sentinel 2A data with our geo-location embedded ground truth. On the other hand, Bing satellite imagery/(Google imagery) is only available in the RGB color profile. Since Bing does not provide other bands, such as near-infrared (NIR) or short-wave infrared (SWIR), standard GIS metrics, such as NDVI and NDWI, cannot be calculated from Bing images.

This paper provides high-resolution pixel-by-pixel Land Use Land Cover (LULC) ground truth information for Dhaka metropolitan city and its surrounding area. This data can serve as LULC ground truth from a developing country. It contains both rural and urban areas. Additionally, several experiments have been performed to establish baseline performance with Bing Data and Sentinel 2A data for five major LULC classes: forest, farmland, builtup, water, and meadow. As Sentinel 2A provides multi-channel information, index images and different combination images are also used for our experiments on LULC segmentation. Finally, data level and decision level fusions are used in the experiments to combine additional information at different levels. The investigation suggests that Bing data (2.22 meter/pixel) consistently outperformed the sentinel 2A data with any forms with a significant margin. However, suppose the high-resolution image is not available. In that case, short wave infrared (SWI) from sentinel data should be used to get the best performance with the deep learning models.

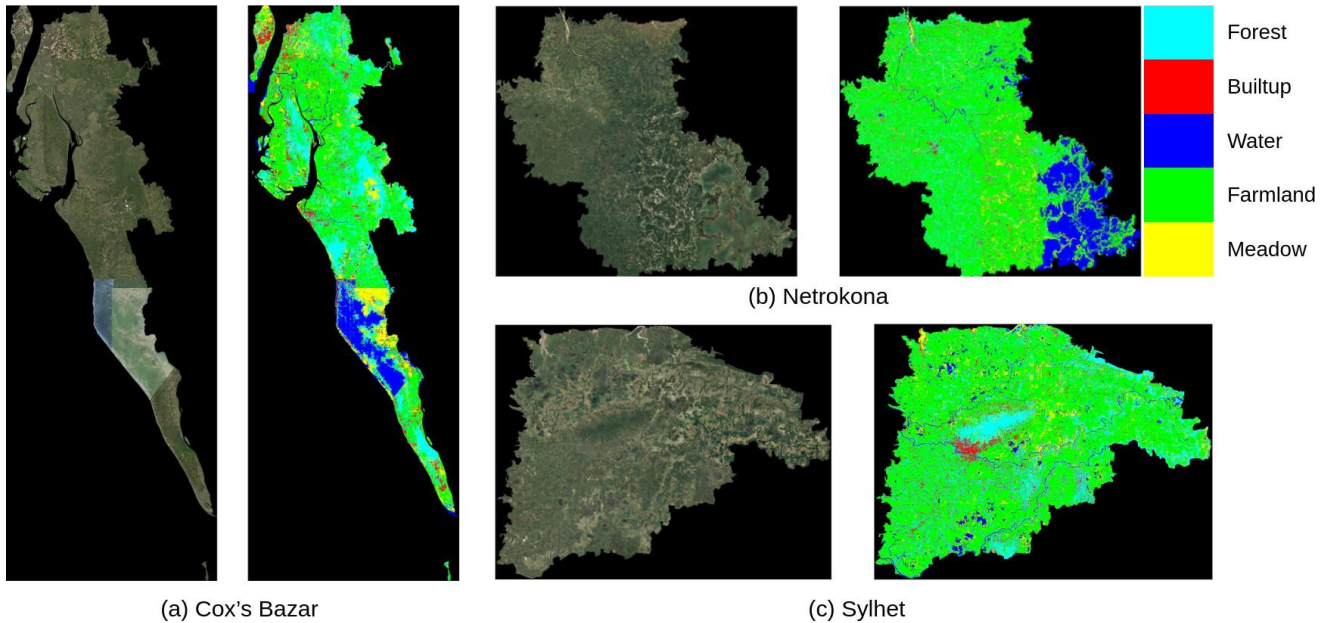

**Figure 14.** Prediction of (a) Cox's Bazar, (b) Netrokona and (c) Sylhet

## Data availability

This dataset has been made available online here under the Creative Common (CC) license: https://doi.org/10.7910/DVN/LLR3RR[61]. Please note that author names are removed to maintain the anonymity of the manuscript.

## Code availability

Custom codes were written to process this dataset. The RGB image is required to convert into Index image as it is required in the architecture. Also, to enhance the performance of the model and eliminate possible confusion from the dataset, the unknown part of the label needs to be projected on the input image. If the paper is accepted, link for the codes will be published here.

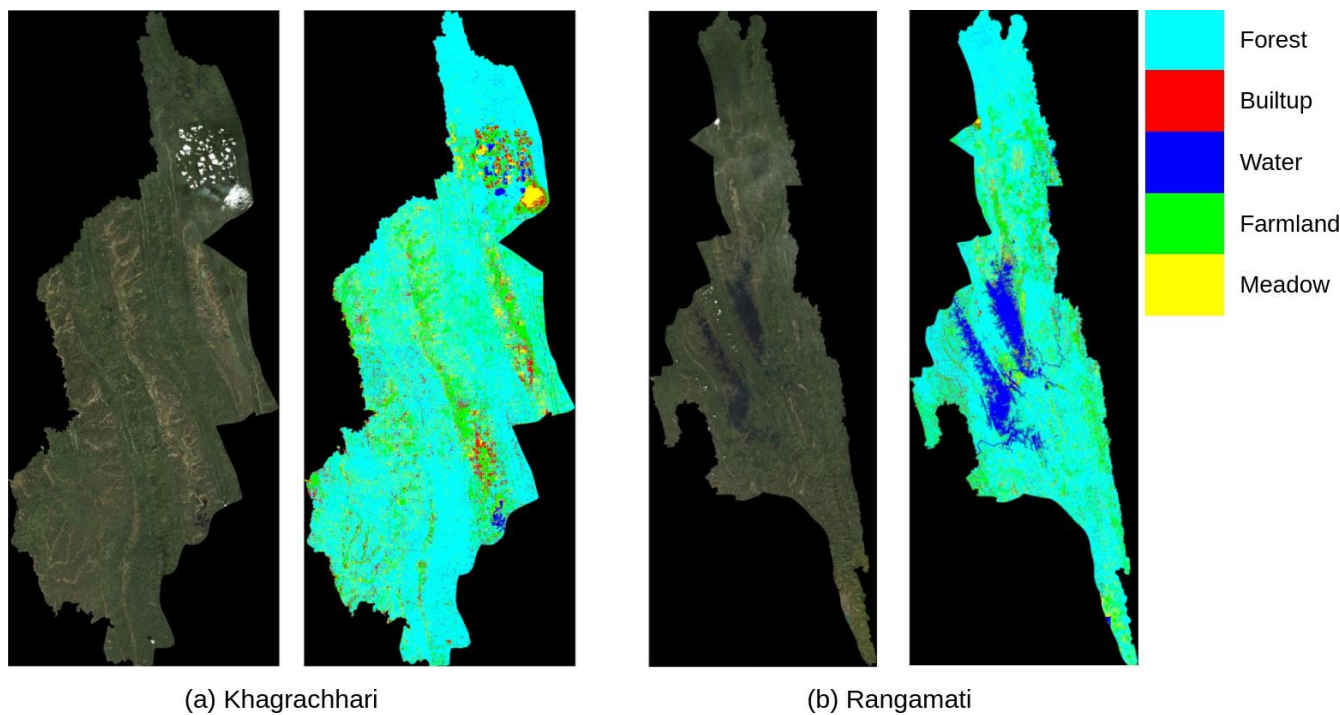

(a) Khagrachhari  (b) Rangamati

**Figure 15.** Prediction of (a) Khagrachhari and (b) Rangamati.

## Author contributions statement

O.P. was the primary author who streamlined the process for BD-SAT data annotation, oversaw the experiments, wrote data pre-processing code, collected the data, created the data pipelines, pre-processed the dataset, modified the Deep Learning architecture and contributed to writing. A.N conducted the experiments, contributed to writing, assisted in data collection, oversaw data annotation and wrote code for data processing. A.S. analyzed the results, result visualization, wrote codes for data processing, assisted in data collection and modified the Deep Learning architecture. A.A.A and M.A.A supported and contributed to technical validation, developing class structure and provided guidance. A.M.R. conducted result analysis, formulated the experiment process, oversaw the experiments, oversaw the annotation process, provided guidance and contributed to writing.

## Competing interests

The authors declare that they have no conflicts of interest.